\documentclass{article}
\usepackage{spconf,amsmath,graphicx}
\usepackage{amssymb}
\usepackage{url}
\usepackage{enumitem}
\setlist{nosep, leftmargin=14pt}

\usepackage{mwe} % to get dummy images
\usepackage{titlesec} % fix for \paragraph undefined control sequence
\usepackage{booktabs}
\usepackage{tabularx}
\usepackage{array}
\usepackage{float}
\usepackage{pifont} % \ding{51} tick, \ding{55} cross

\newcolumntype{Y}{>{\raggedright\arraybackslash}X}
\newcolumntype{C}{>{\centering\arraybackslash}p{1.2cm}} % nhỏ gọn cho F1
\title{Robust White Blood Cell Classification with Stain-Normalized Decoupled Learning and Ensembling}
\name{
\begin{tabular}{c}
Luu Le$^{1,3*} \qquad
$Hoang-Loc Cao$^{2,3*}$\qquad
Ha-Hieu Pham$^{2,3}$\qquad
Thanh-Huy Nguyen$^{4}$\qquad
Ulas Bagci$^{5\dagger}$
\thanks{$\dagger$ Corresponding Author} \thanks{$^{\ast}$ Equal contribution.}
\end{tabular}}
\address{
$^{1}$University of Technology, Ho Chi Minh City, Vietnam\\ 
$^{2}$ University of Science, Ho Chi Minh City, Vietnam \\
$^{3}$ Vietnam National University, Ho Chi Minh City, Vietnam \\
$^{4}$ Carnegie Mellon University, Pittsburgh, PA, USA \\
$^{5}$ Northwestern University, Chicago, IL, USA}
\begin{document}
%\ninept
%
 \maketitle
\begin{abstract}
White blood cell (WBC) classification is fundamental for hematology applications such as infection assessment, leukemia screening, and treatment monitoring. However, real-world WBC datasets present substantial appearance variations caused by staining and scanning conditions, as well as severe class imbalance in which common cell types dominate while rare but clinically important categories are underrepresented. To address these challenges, we propose a stain-normalized, decoupled training framework that first learns transferable representations using instance-balanced sampling, and then rebalances the classifier with class-aware sampling and a hybrid loss combining effective-number weighting and focal modulation. In inference stage, we further enhance robustness by ensembling various trained backbones with test-time augmentation. Our approach achieved the top rank on the leaderboard of the WBCBench 2026: Robust White Blood Cell Classification Challenge at ISBI 2026.

\end{abstract}

\section{Introduction}
\label{sec:intro}

Automating the analysis of peripheral blood smear images has become an objective in digital pathology. Detection and classification of White Blood Cells (WBCs) are essential for monitoring immune health and diagnosing hematological disorders. However, manual microscopy, the current gold standard, is time consuming and subjective, motivating the development of computer aided diagnostic systems. Deep learning has advanced WBC analysis through convolutional neural networks (CNNs) and, more recently, Vision Transformers (ViTs), enabling automatic extraction of discriminative features without labor-intensive preprocessing~\cite{girdhar2022classification,zhu2026bcct}. Despite these advances, deploying automated systems remains challenging. A major issue is domain shift arising from variations in staining protocols, imaging devices, and acquisition settings. Differences in color distribution, contrast, resolution, and cellular morphology can degrade performance on unseen domains. Moreover, WBC datasets are typically long-tailed, where common cell types dominate and rare subtypes are scarce. This imbalance biases models toward majority classes and reduces sensitivity to minority categories.

% The WBCBench 2026 Robust White Blood Cell Classification Challenge aims to benchmark robustness under these conditions that emphasize cross-domain generalization and long-tailed recognition without labeled target domain data. The evaluation protocol prioritizes macro-averaged and balanced metrics over accuracy, highlighting the need for transferable representations that remain stable under distribution shifts while maintaining per class fairness. Previous works addressed these issues using explicit domain adaptation strategies. Chossegros~\cite{chossegros2024improving} proposed a few-shot domain adaptation framework to improve cross-source generalization. Li~\cite{li2024stain} mitigated domain shift and imbalance via stain-based augmentation and local alignment with a decoupled two-stage scheme. Gao~\cite{gao2025danet} introduced DaNet, combining balanced multi-domain mixup with distribution alignment. However, such approaches often rely on explicit alignment objectives, increasing training complexity.

The WBCBench 2026 Robust White Blood Cell Classification Challenge benchmarks robustness under cross-domain generalization and long-tailed recognition without labeled target data. The evaluation protocol prioritizes macro-averaged and balanced metrics over accuracy, emphasizing transferable representations that remain stable under distribution shifts while maintaining per-class fairness. Previous works addressed these challenges using explicit domain adaptation. Chossegros~\cite{chossegros2024improving} proposed few-shot adaptation for cross-source generalization. Li~\cite{li2024stain} mitigated domain shift and imbalance via stain augmentation and local alignment in a two-stage scheme. Gao~\cite{gao2025danet} introduced DaNet, combining multi-domain mixup with distribution alignment. However, these methods rely on alignment objectives, increasing training complexity.

In contrast, decoupled long-tailed learning first learns representations under the natural distribution, then recalibrates the classifier separately~\cite{kang2019decoupling,pham2026handling}. We propose a decoupled framework for long-tailed WBC classification: applying Macenko normalization~\cite{macenko2009method}, instance-balanced sampling for representation learning, class-aware rebalancing for classifier calibration, and enhancing robustness via test-time augmentation and backbone ensembling.

\begin{figure*}[ht]
  \centering
  \includegraphics[width=1.0\linewidth]{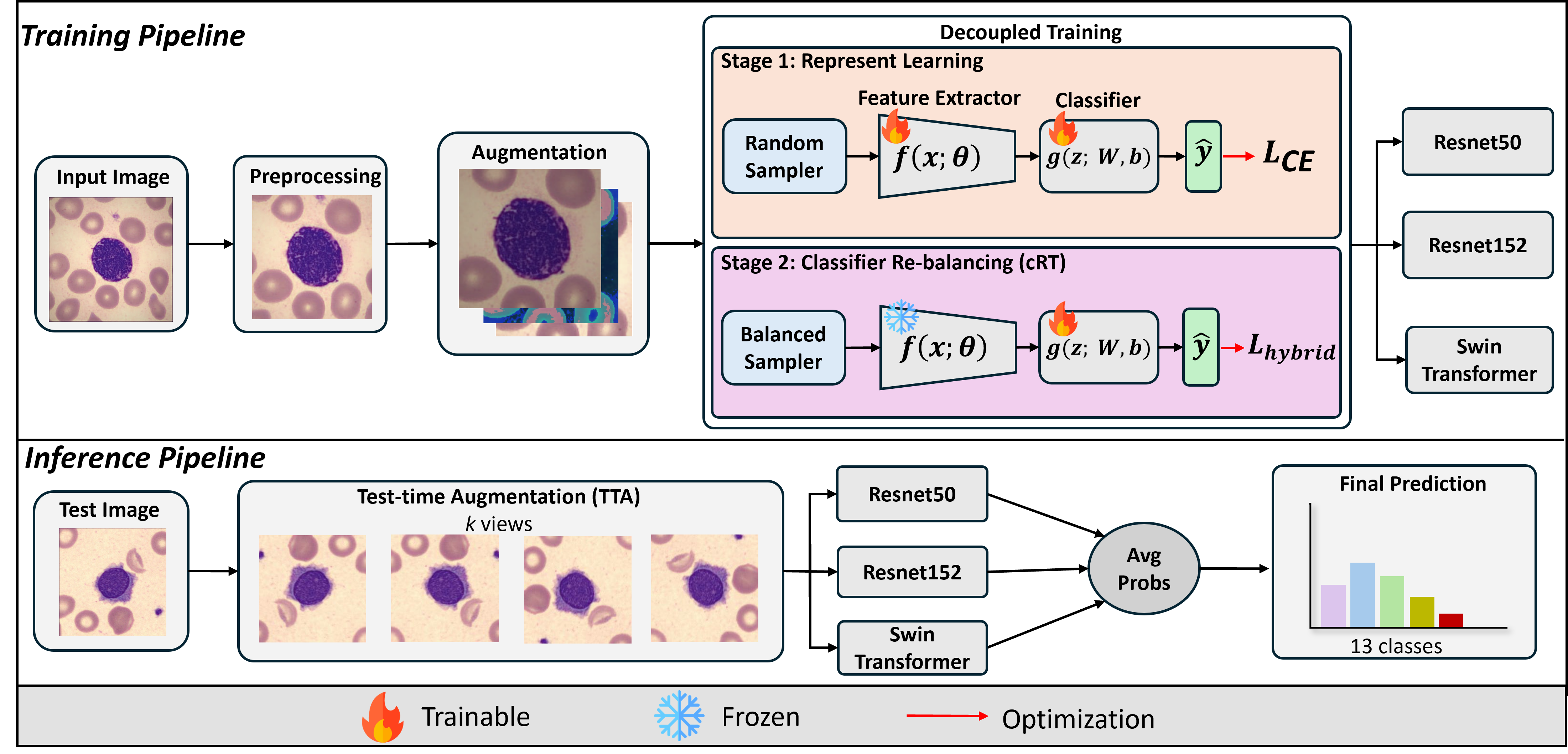}
  \caption{Overview of our proposed framework for WBC classification}
  \label{fig:framework}
\end{figure*}

\section{Methodology}
\label{sec:method}

Fig.~\ref{fig:framework} illustrates our proposed framework for robust white blood cell classification under long tailed distribution and cross domain shift.

\subsection{Data Preprocessing}

Histopathology images exhibit substantial color variation due to differences in staining protocols, scanners, and acquisition settings. To reduce this domain discrepancy, we apply Macenko stain normalization, which converts RGB images into optical density (OD) space, estimates dominant stain vectors via singular value decomposition, and projects each image onto a standardized stain template. This procedure aligns stain appearance across domains while preserving structural morphology. After normalization, images are resized to $368 \times 368$ pixels and center cropped to $224 \times 224$. During training, we apply RandAugment~\cite{cubuk2020randaugment} with random geometric and photometric transformations to enhance generalization and simulate potential domain variations. All pixel values are normalized using ImageNet statistics to ensure compatibility with pretrained backbones.

\subsection{Decoupled Two Stage Training}

Given a training dataset $\mathcal{D} = \{(x_i, y_i)\}_{i=1}^{n}$ with $C$ classes and class frequencies $\{n_j\}_{j=1}^{C}$, our model consists of a feature extractor $f(x;\theta)$ and a linear classifier $g(z;W,b)=W^\top z+b$, where $z=f(x;\theta)$.

To alleviate representation bias induced by long tailed distributions, we adopt a decoupled training strategy that separates feature learning from classifier re balancing.

\subsubsection{Stage 1: Representation Learning}

In Stage 1, we train the entire network end to end using instance balanced (random) sampling, where mini batches follow the empirical data distribution. The sampling probability of class $j$ is

\begin{equation}
p_j = \frac{n_j}{\sum_{i=1}^{C} n_i}.
\end{equation}
Sampling is performed without replacement within each epoch so that every sample is visited once. Both backbone parameters $\theta$ and classifier parameters $(W,b)$ are optimized using standard Cross Entropy loss. This stage aims to learn transferable representations that capture intrinsic morphological features without explicitly altering the data prior.

\subsubsection{Stage 2: Classifier Re balancing (cRT)}

After Stage 1, the learned classifier is biased toward head classes. In Stage 2, we freeze the backbone $f(x;\theta)$, re initialize the classifier, and retrain only $(W,b)$ on fixed embeddings $z$ to recalibrate decision boundaries.

\paragraph{Class balanced sampling.}
We adopt inverse frequency sampling with weight $w_i = \frac{1}{n_{y_i}}$, and sample with replacement to approximate a uniform class prior across mini batches.

\paragraph{Hybrid class balanced loss.}
To further compensate for imbalance, we apply effective number re-weighting~\cite{cui2019class}. For class $j$,

\begin{equation}
E_{n_j}=\frac{1-\beta^{n_j}}{1-\beta}, \quad
\alpha_j=\frac{1-\beta}{1-\beta^{n_j}}.
\label{eq:effective}
\end{equation}
For a sample $(x,y)$ with predicted probability $p_y$, the class balanced cross entropy is
\begin{equation}
\mathcal{L}_{CE} = -\alpha_y \log p_y.
\label{eq:lossce}
\end{equation}
To emphasize hard examples, we incorporate focal modulation with focusing parameter $\gamma$:
\begin{equation}
\mathcal{L}_{Focal} = (1-p_y)^{\gamma} \mathcal{L}_{CE}.
\label{eq:loss}
\end{equation}
The final Stage 2 objective is a convex combination:
\begin{equation}
\mathcal{L}_{Stage2} = (1-\lambda)\mathcal{L}_{CE}
+ \lambda \mathcal{L}_{Focal}.
\label{eq:loss_hybrid}
\end{equation}
This hybrid objective stabilizes optimization while improving minority class sensitivity.

\subsection{Inference with Test Time Augmentation and Ensembling}

We three backbones with complementary inductive biases: two CNNs (ResNet50 and ResNet152) and one Transformer (Swin Transformer), all initialized from ImageNet-pretrained weights. During inference, each test image is processed using the same stain normalization and resizing pipeline as in training, and then evaluated with TTA. 
Specifically, for an ensemble of $M$ models (indexed by $m\in\mathcal{M}$), we generate $K$ augmented views per image and obtain the corresponding softmax probability vectors $\mathbf{p}^{(m)}_{k}$. The final prediction distribution is computed by averaging probabilities over both models and augmentations:
\begin{equation}
\mathbf{p}_{\mathrm{ens}}
= \frac{1}{M K}\sum_{m\in\mathcal{M}}\sum_{k=1}^{K}\mathbf{p}^{(m)}_{k}.
\label{eq:tta_ens}
\end{equation}
The predicted class is then given by
\begin{equation}
\hat{y} = \arg\max_{c}\mathbf{p}_{\mathrm{ens}}[c].
\label{eq:pred}
\end{equation}
This combined strategy reduces prediction variance and improves robustness to cross-domain shifts, without requiring any explicit target-domain adaptation.

\section{Experiments}
\label{sec:exp}
\subsection{Implementation Details and Baselines}
All experiments were conducted on Kaggle using dual Tesla T4 GPUs (16 GB VRAM each). We trained three models: ResNet50 (R50), ResNet152 (R152) and Swin Transformer (Swin) with mixed precision (AMP) and a batch size of 256 using a two-stage strategy: first training the backbone for 100 epochs with AdamW (lr=$1\times10^{-4}$) under random sampling, then retraining the classifier for 50 epochs with class-balanced sampling and AdamW (lr=$1\times10^{-3}$). A cosine annealing scheduler and early stopping were applied in both stages. In Stage 2, we used effective-number class weighting with $\beta=0.9999$ and a combined weighted Cross-Entropy and Focal loss ($\gamma=2.0$, $\lambda=0.5$). During inference, we employed test-time augmentation with $K=8$ views and ensembled the two models by averaging their predicted probabilities.

We evaluate our method on the official WBCBench 2026 Challenge benchmark~\cite{wbcbench2026}. We first reserve 20\% of the entire dataset as a held-out test set. 
The remaining 80\% is used to perform five-fold cross-validation for model training and validation. 
Final performance is reported on the held-out test set. We compare (i) standard end-to-end supervised training on each backbone (R50/R152/Swin) and (ii) a two-stage decoupled variant that retrains the classifier with class-balanced sampling. We also report probability-averaging ensembles (Ens.) of different backbones to assess robustness under cross-domain shift. Performance is measured using Macro-F1, Balanced Accuracy, Macro-Precision, and Macro-Specificity to address class imbalance.

\begin{table}[H]
\centering
\caption{Main comparison on the WBCBench 2026 Challenge benchmark. Best results are in \textbf{bold}; second-best are \underline{underlined}.}
\label{tab:main_results}
\small
\setlength{\tabcolsep}{3pt}
\renewcommand{\arraystretch}{1.15}

\begin{tabularx}{\columnwidth}{l CCCC}
\toprule
\textbf{Method} 
& \textbf{MF1}$\uparrow$ 
& \textbf{BAcc}$\uparrow$ 
& \textbf{MP}$\uparrow$ 
& \textbf{MSpec}$\uparrow$ \\
\midrule

Supervised (R50)   & 65.8±0.6 & 63.2±0.9 & 71.2±0.8 & \underline{99.3±0.0} \\
Supervised (R152)  & 68.7±2.2 & 66.1±1.1 & \underline{75.5±5.4} & \textbf{99.4±0.0}\\
Supervised (Swin)  & 71.2±1.4 & 69.0±1.3 & \textbf{76.4±3.8} & \textbf{99.4±0.0}\\
Decoupled (R50)    & 70.4±2.6 & 72.0±2.3 & 70.0±3.4 & \underline{99.3±0.0} \\
Decoupled (R152)   & 70.6±2.1 & 70.9±1.4 & 71.9±4.0 & \textbf{99.4±0.0} \\
Decoupled (Swin)   & 66.0±2.2 & \textbf{80.9±1.2} & 61.9±2.1 & 99.2±0.1 \\

\midrule
Ens. (R50+R152)     & \underline{72.4±2.6} & 72.7±2.1 & 73.0±2.9 & \textbf{99.4±0.0} \\
Ens. (R50+Swin)     & 72.0±1.4 & \underline{79.1±1.1} & 69.0±2.0 & \textbf{99.4±0.0} \\
Ens. (R152+Swin)    & 72.3±0.8 & 79.0±1.7 & 69.2±1.1 & \textbf{99.4±0.0} \\
Ens. (all)          & \textbf{74.2±1.0} & 77.1±0.7 
                    & 72.6±1.8 & \textbf{99.4±0.0} \\
\bottomrule
\end{tabularx}
\end{table}

\subsection{Results and Analysis}

Table~\ref{tab:main_results} shows that decoupled learning consistently improves class-balanced performance, as reflected by higher BAcc for both CNN backbones. R50 increases from 63.2 to 72.0 and R152 from 66.1 to 70.9, indicating reduced bias toward majority classes. MF1 also improves over supervised baselines, while MP gains are less consistent, suggesting a trade-off between enhanced minority recall and precision stability. Although Swin achieves the highest single-model BAcc (80.9) under decoupling, its lower MF1 and MP indicate different calibration behavior across architectures.

\begin{figure}[t]
  \centering
  \includegraphics[width=\columnwidth]{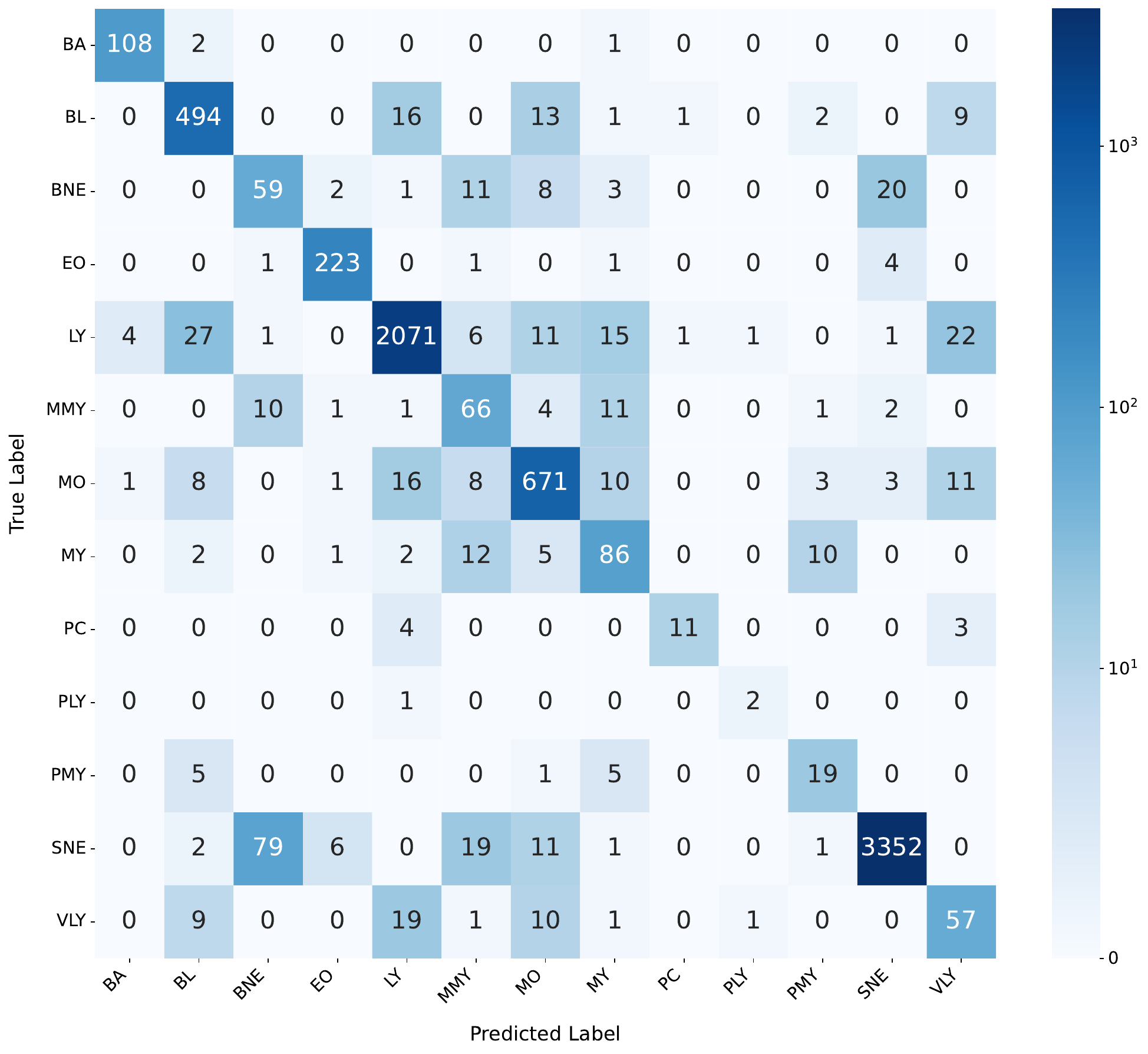}
  \caption{Confusion matrix of our best model on WBCBench 2026 Challenge Benchmark.}
  \label{fig:confusion}
\end{figure}

% \begin{table}[H]
% \centering
% \caption{Leaderboard on the test set (Prize Contenders).}
% \label{tab:leaderboard}
% \small
% \setlength{\tabcolsep}{6pt}
% \renewcommand{\arraystretch}{1.1}

% \begin{tabular}{c l c}
% \toprule
% \textbf{\#} & \textbf{Team} & \textbf{Macro-Averaged F1}$\uparrow$ \\
% \midrule
% 1  & FDVTS\_WBC        & 0.75661 \\
% 2  & PathMedAI         & 0.75401 \\
% 3  & jht010312         & 0.74013 \\
% 4  & CPRL              & 0.72035 \\
% 5  & PACV              & 0.71875 \\
% \textbf{6}  & \textbf{AIO-MHIL} & \textbf{0.70817} \\
% 7  & Quan H. Cap       & 0.70368 \\
% 8  & GODA              & 0.68586 \\
% 9  & ayatullah Elady98 & 0.68164 \\
% 10 & smart\_lab        & 0.68024 \\
% \bottomrule
% \end{tabular}
% \end{table}

Ensembling further enhances robustness. Ens.\ (R50 \& R152) raises MF1 to 72.4, while the full ensemble achieves the best overall performance (MF1 = 74.2, BAcc = 77.1), confirming the complementary strengths of heterogeneous models under domain shift.

Fig.~\ref{fig:confusion} highlights improved minority recognition. Despite scarce samples, rare classes such as PLY (2/3) and PC (11/18) are correctly identified, and other underrepresented categories (e.g., PMY, MMY) attain reasonable true positive rates. Nonetheless, the long-tailed setting remains challenging, with some minority instances misclassified into visually similar or dominant classes (e.g., PC $\rightarrow$ LY). Overall, these results demonstrate that our class-aware rebalancing and hybrid loss enhance minority sensitivity without substantially compromising majority performance.

\section{Conclusion}

In this work, we tackled long-tailed WBC classification under cross-domain shifts in the WBCBench 2026 challenge. Our method combines decoupled training for representation and classifier recalibration with stain normalization, class-aware rebalancing, TTA, and multi-backbone ensembling. Experiments show consistent gains in class-balanced performance and improved robustness across domains. The final framework achieves competitive test results without using labeled target-domain data, highlighting the effectiveness of simple, well-structured training and inference strategies for realistic long-tailed medical imaging settings.

% \section{Acknowledgments}
% \label{sec:acknowledgments}

\bibliographystyle{IEEEbib}
\bibliography{strings,refs}

\end{document}